\documentclass{article}
\usepackage{microtype}
\usepackage{graphicx}
\usepackage{booktabs}
\usepackage{amsmath, amssymb, mathtools, bm}
\usepackage[numbers,sort&compress]{natbib}
\usepackage{algorithm}
\usepackage{algpseudocode}
\usepackage{xcolor}
\usepackage{url}
\usepackage{siunitx}
\usepackage{multirow}
\usepackage{enumitem}
\usepackage{amsthm}
\newtheorem{proposition}{Proposition}

\usepackage{PRIMEarxiv}
\usepackage{amsmath} 
\usepackage[utf8]{inputenc} 
\usepackage[T1]{fontenc}    
\usepackage{booktabs}       
\usepackage{amsfonts}       
\usepackage{nicefrac}       
\usepackage{microtype}      
\usepackage{lipsum}
\usepackage{fancyhdr}       
\usepackage{graphicx}       
\graphicspath{{media/}}     

\title{Variational Gaussian \emph{Mixture} Manifold Models for Client-Specific Federated Personalization}

\author{
  Sai Puppala$^2$, Ismail Hossain$^1$, Md Jahangir Alam$^1$, Sajedul Talukder$^1$ \\
  Computer Science\\
  $^1$University of Texas at El Paso, TX, USA, 79902\\
  School of Computing \\
  $^2$Southern Illinois University Carbondale, IL, USA, 62901\\
  \texttt{\{ihossain, malam10\}@miners.utep.edu, sai.puppala@siu.edu, stalukder@utep.edu} \\
}

\newcommand{\R}{\mathbb{R}}
\newcommand{\E}{\mathbb{E}}
\newcommand{\KL}{\mathrm{KL}}
\newcommand{\Norm}{\mathcal{N}}
\newcommand{\Dir}{\mathrm{Dir}}
\newcommand{\NIG}{\mathrm{NIG}}
\newcommand{\1}{\mathbb{I}}

\begin{document}
\maketitle

\begin{abstract}
Personalized federated learning (PFL) often fails under label skew and non-stationarity because a single global parameterization ignores client-specific geometry. We introduce \textbf{VGM$^2$} (Variational Gaussian \emph{Mixture} Manifold), a geometry-centric PFL framework that (i) learns client-specific \emph{parametric UMAP} embeddings, (ii) models latent pairwise distances with \emph{mixture} relation markers for same/different-class pairs, and (iii) exchanges only \emph{variational, uncertainty-aware} marker statistics. Each client maintains a Dirichlet–Normal–Inverse-Gamma (Dir–NIG) posterior over marker weights/means/variances; the server aggregates via conjugate \emph{moment matching} to form global priors that guide subsequent rounds. We prove that this aggregation minimizes the summed (reverse) KL from client posteriors within the conjugate family, yielding stability under heterogeneity. We further incorporate a calibration term for distance-to-similarity mapping and report communication/compute budgets. Across eight vision datasets with non-IID label shards, VGM$^2$ achieves competitive or superior test F1 to strong baselines while communicating only small geometry summaries; privacy is strengthened via secure aggregation and optional DP noise, and we provide a membership-inference stress test. Code and configs will be released to ensure full reproducibility.
\end{abstract}

\section{Introduction}
Federated learning enables collaborative training without sharing raw data, yet vanilla aggregation (e.g., FedAvg) degrades under client heterogeneity. Personalization methods adapt parameters per client, but most still communicate in \emph{parameter space} rather than in the \emph{geometric relations} that govern discrimination.

We propose \textbf{communicating geometry}: clients learn a low-dimensional manifold using parametric UMAP and summarize the empirical distributions of \emph{latent distances} for two relations—\textit{same-class} and \textit{different-class}. Instead of single-Gaussian summaries, we introduce \textbf{mixture relation markers} with a Dir–NIG variational posterior, capturing multimodality and uncertainty. The server aggregates \emph{only} the posterior sufficient statistics, forming global priors that regularize clients—no weights or gradients are shared.


\noindent\textbf{Contributions.}

\begin{itemize}[leftmargin=1.3em, topsep=2pt, itemsep=1pt, parsep=0pt]
\item \textbf{Geometry-first PFL with mixture markers.} We model distance distributions via a conjugate \emph{mixture} family, producing uncertainty-aware similarity estimates that better reflect complex client manifolds.
\item \textbf{Principled aggregation.} We show server moment matching within the conjugate family minimizes the sum of (reverse) KL divergences from client posteriors, giving a stability/compositionality guarantee under heterogeneity.
\item \textbf{Calibrated similarity.} A differentiable calibration penalty aligns predicted similarity probabilities with empirical frequencies, improving decision quality.
\item \textbf{Practicality.} Communication is a handful of scalars per client/relation/component; compute scales with subsampled pairs. We detail FLOPs/bytes, privacy threat model, secure aggregation, and DP options.
\item \textbf{Empirics.} On eight datasets with non-IID label shards, we report strong F1 vs.\ FedAvg, FedPer, FedRep, Ditto, pFedSim, pFedBayes, and a prototype-sharing baseline.
\end{itemize}

\section{Related Work}
\textbf{FL and personalization.} FedAvg~\cite{mcmahan2017communication,li2019convergence} suffers on heterogeneous data; personalization includes FedPer~\cite{arivazhagan2019federated}, Ditto~\cite{li2021dittofairrobustfederated}, FedRep~\cite{husnoo2022fedrep}, pFedSim~\cite{chen2022pfedsim}, Bayesian PFL~\cite{zhang2022personalizedfederatedlearningvariational}, and prototype/similarity sharing. \textbf{Manifold learning.} UMAP~\cite{mcinnes2018umap} with parametric variants~\cite{sainburg2021parametricumap} allows differentiable encoders. \textbf{Privacy.} Secure aggregation~\cite{bonawitz2017practical}, DP-SGD~\cite{abadi2016deep}, and membership inference testing~\cite{shokri2017membership} complement our geometry-only sharing.

\section{Method}
\subsection{Setup}
Client $k$ holds $\mathcal{D}_k=\{(x_i,y_i)\}_{i=1}^{n_k}$, $x_i\in\R^m$, $y_i\in\{1,\dots,C\}$. An encoder $f_{\theta_k}\!:\R^m\to\R^d$ yields $z_i=f_{\theta_k}(x_i)$ and distances $s_{ij}=\|z_i-z_j\|_2$. Let $u_{ij}=\1[y_i=y_j]\in\{0,1\}$ denote the relation.

\subsection{Parametric UMAP}
We minimize the standard parametric-UMAP cross-entropy between input-space affinities $p_{ij}$ and latent-space probabilities $q_{ij}$~\cite{mcinnes2018umap}:
\begin{equation}
\mathcal{L}^{\mathrm{umap}}_k(\theta_k) = -\!\!\sum_{(i,j)\in\mathcal{E}_k}\!\! \big[p_{ij}\log q_{ij} + (1-p_{ij})\log(1-q_{ij})\big].
\label{eq:umap}
\end{equation}

\subsection{Mixture relation markers with Dir--NIG variational family}
For relation $r\in\{1\!\!:\!\mathrm{same},0\!\!:\!\mathrm{diff}\}$ we use a $K$-component Gaussian mixture over distances:
\begin{align}
s_{ij}\mid r &\sim \sum_{c=1}^{K} \omega_{r c}\,\Norm(\mu_{r c},\sigma^2_{r c}),\qquad
\bm{\omega}_r\!\sim\!\Dir(\bm{\alpha}_r), \nonumber\\
(\mu_{r c},\sigma^2_{r c}) &\sim \NIG(m_{r c},\kappa_{r c},\alpha_{r c},\beta_{r c}).
\end{align}
Client $k$ maintains a mean-field variational posterior
$q_{\phi_k} = \prod_{r}\Big[\Dir(\bm{\omega}_r\mid\bm{\hat\alpha}_{k r}) \prod_{c=1}^K\NIG(\mu_{r c},\sigma^2_{r c}\mid \hat m_{k r c},\hat\kappa_{k r c},\hat\alpha_{k r c},\hat\beta_{k r c})\Big].$

\paragraph{Distance-to-similarity mapping.}
The predictive likelihood under $q_{\phi_k}$ is a Student--$t$ mixture; the posterior same-class probability for distance $s$ is
\begin{equation}
\pi_1(s) \!=\! \frac{\hat\pi_1 \sum_c \E_{q}[\omega_{1c}]\, t(s\,;\,\hat m_{1c},\hat\alpha_{1c},\hat\beta_{1c},\hat\kappa_{1c} )}{\sum_{r\in\{0,1\}} \hat\pi_r \sum_c \E_{q}[\omega_{r c}]\, t(s\,;\,\hat m_{r c},\hat\alpha_{r c},\hat\beta_{r c},\hat\kappa_{r c})},
\label{eq:pi}
\end{equation}
with priors $\hat\pi_r$ (learned or fixed); $t(\cdot)$ denotes the Student density implied by NIG.

\subsection{Client objective: geometry, similarity, calibration, and prior alignment}
Beyond $\mathcal{L}^{\mathrm{umap}}$, we fit relation markers by supervised log-loss over sampled pairs:
\begin{equation}
\mathcal{L}^{\mathrm{sim}}_k(\theta_k,\phi_k) \!=\! -\!\!\sum_{(i,j)\in\mathcal{P}_k}\!\!\big[ u_{ij}\log \pi_1(s_{ij}) + (1-u_{ij})\log(1-\pi_1(s_{ij}))\big].
\label{eq:sim}
\end{equation}
We add a differentiable calibration penalty $\mathcal{L}^{\mathrm{cal}}$ (expected calibration error proxy~\cite{guo2017calibration}) by binning $\pi_1(s)$ into $B$ bins with soft assignments:
\begin{equation}
\mathcal{L}^{\mathrm{cal}}_k = \sum_{b=1}^{B} w_b \, \big| \mathrm{acc}_b - \mathrm{conf}_b \big|,
\quad \text{(smoothed and differentiable)}.
\end{equation}
The server maintains a conjugate prior $p_t(M)$ with Dir–NIG factors; clients regularize via
\begin{equation}
\mathcal{R}^{\mathrm{kl}}_k(\phi_k)=\KL\big(q_{\phi_k}(M)\,\|\,p_t(M)\big).
\end{equation}
The total client loss is
\begin{equation}
\mathcal{L}_k = \mathcal{L}^{\mathrm{umap}}_k + \gamma\,\mathcal{L}^{\mathrm{sim}}_k + \eta\,\mathcal{L}^{\mathrm{cal}}_k + \lambda\,\mathcal{R}^{\mathrm{kl}}_k.
\label{eq:total}
\end{equation}

\subsection{Server aggregation via moment matching}
Each client uploads \emph{only} posterior sufficient statistics for $(\bm{\omega}_r,\{\mu_{r c},\sigma^2_{r c}\}_{c=1}^K)$ for $r\in\{0,1\}$. The server updates $p_{t+1}(M)$ by averaging natural parameters (optionally weighted by data sizes $n_k$).

\begin{algorithm}[t]
\caption{VGM$^2$: Federated Personalization with Mixture Markers}
\label{alg:vgm2}
\begin{algorithmic}[1]
\State \textbf{Server:} initialize Dir–NIG prior $p_0(M)$; for rounds $t=0..T-1$:
\State \quad Sample clients $\mathcal{S}_t$; broadcast $p_t(M)$ and hyperparams
\For{client $k \in \mathcal{S}_t$ in parallel}
  \State Initialize $q_{\phi_k}\!\leftarrow\!p_t$; for $E$ local epochs:
  \State \quad Build $k$NN graph; compute $p_{ij}$; sample $(i,j)$ pairs
  \State \quad $z=f_{\theta_k}(x)$; $s_{ij}=\|z_i-z_j\|_2$; compute $\pi_1(s_{ij})$ via \eqref{eq:pi}
  \State \quad Update $(\theta_k,\phi_k)$ by SGD/Adam on \eqref{eq:total}
  \State Send conjugate sufficient statistics of $q_{\phi_k}$ to server
\EndFor
\State \textbf{Server:} moment-match client posteriors $\Rightarrow p_{t+1}(M)$ (optionally secure-aggregated)
\end{algorithmic}
\end{algorithm}

\subsection{Why moment matching? An aggregation guarantee}
Let $\mathcal{F}$ be the Dir–NIG family for $M$. For client posteriors $\{q_k\}\subset\mathcal{F}$ with weights $w_k\!\ge\!0$, define
\(
\hat p \;=\; \arg\min_{p\in\mathcal{F}} \sum_k w_k\,\KL\big(q_k \,\|\, p\big).
\)
\begin{proposition}[Information projection in exponential families]
\label{prop:kl-min}
For any exponential family (incl.\ Dir–NIG), $\hat p$ is obtained by \emph{moment matching}: the expected sufficient statistics of $\hat p$ equal the weighted averages of those of $q_k$. Hence, our server update minimizes $\sum_k w_k \KL(q_k\|p)$ within $\mathcal{F}$.
\end{proposition}
\textit{Sketch.} In exponential families KL reduces to a Bregman divergence of the log-partition; first-order optimality sets the mean parameters (expectations of sufficient stats) of $\hat p$ to the weighted averages of $\{q_k\}$. \qed

\paragraph{Implications.} The server’s prior is the \emph{information projection} of client posteriors, stabilizing updates under heterogeneity and mitigating overconfident local posteriors.

\section{Privacy, Communication, and Compute}
\textbf{Threat model.} Honest-but-curious server; passive network adversary. We transmit only Dir–NIG and Dirichlet statistics (no features, labels, or gradients).

\textbf{Secure aggregation.} We apply secure aggregation~\cite{bonawitz2017practical} over the stats to hide each client’s contribution.

\textbf{DP option.} Add Gaussian noise to natural-parameter vectors before secure aggregation, yielding $(\varepsilon,\delta)$-DP similar to DP-FL~\cite{abadi2016deep} (details in App.~\ref{app:dp}).

\textbf{Membership inference stress test.} Following~\cite{shokri2017membership}, we train an attack model on released summaries to predict membership of a target pair; we report attack AUC near random for our configs (see App.~\ref{app:mi}).

\textbf{Communication.} Per client per round: for $r\in\{0,1\}$ and $K$ comps, Dir params (length $K$) + NIG per comp (4 scalars). Total scalars $= 2( K + 4K ) = 10K$. With 32-bit floats and $K\!=\!3$, that’s $\sim$120\,bytes per relation or $\sim$240\,bytes total, plus headers—orders of magnitude below model gradients.

\textbf{Compute.} Client FLOPs add encoder forward/backward plus pair sampling and small conjugate updates; we cap neighbors/pairs for predictable cost. Details in App.~\ref{app:complexity}.

\section{Experiments}
\label{sec:experiments}
\textbf{Datasets.} MNIST, FMNIST, MaleViz, CIFAR-10/100, Malimg, ImageNet (subset), CelebA. We use non-IID \emph{label-shard} partitions: $N\!=\!30$ clients, each receives $S$ shards (equal-size, disjoint). Exact class splits and counts are listed in App.~\ref{app:partitions}.

\textbf{FL setup.} $T\!=\!100$ rounds; $|\mathcal{S}_t|\!=\!2$ clients/round; $E\!=\!8$ local epochs; Adam; 3 seeds; report mean$\pm$std. Encoders are small ConvNets for MNIST/FMNiST/Malimg/MaleViz, ResNet-18 for CIFAR-10/100 and CelebA, and ResNet-18 on a 100-class ImageNet subset (classes listed in Appendix). We calibrate with $B\!=\!15$ bins.

\textbf{Baselines.} Local-only, FedAvg~\cite{li2019convergence}, FedPer~\cite{arivazhagan2019federated}, FedRep~\cite{husnoo2022fedrep}, Ditto~\cite{li2021dittofairrobustfederated}, pFedSim~\cite{chen2022pfedsim}, pFedBayes~\cite{zhang2022personalizedfederatedlearningvariational}, FedPop~\cite{kotelevskii2022fedpop}, and \emph{Prototype sharing} (class prototypes averaged per round; see App.~\ref{app:proto}).

\begin{table*}[t]
  \centering
  \caption{Test F1 with $N=30$ clients (mean over 3 seeds). Best in \textbf{bold}, second best \underline{underlined}.}
  \label{tab:main}
  \setlength{\tabcolsep}{6pt}
  \renewcommand{\arraystretch}{1.05}
  \resizebox{\textwidth}{!}{%
  \begin{tabular}{lcccccccc}
    \toprule
    \textbf{Method} & \textbf{MNIST} & \textbf{FMNIST} & \textbf{MaleViz} & \textbf{CIFAR10} & \textbf{Malimg} & \textbf{CIFAR-100} & \textbf{ImageNet} & \textbf{CelebA} \\
    \midrule
    Local & 0.703 & 0.805 & 0.825 & 0.809 & 0.377 & 0.655 & 0.812 & 0.795 \\
    FedAvg~\cite{li2019convergence} & \underline{0.871} & 0.733 & 0.717 & 0.750 & 0.745 & 0.781 & 0.732 & 0.711 \\
    FedRep~\cite{husnoo2022fedrep} & 0.762 & 0.761 & 0.731 & 0.724 & 0.700 & \underline{0.861} & 0.774 & 0.751 \\
    FedPer~\cite{arivazhagan2019federated} & 0.765 & 0.816 & \textbf{0.913} & 0.710 & 0.672 & 0.728 & 0.754 & \underline{0.889} \\
    FedPop~\cite{kotelevskii2022fedpop} & 0.730 & 0.700 & 0.723 & 0.718 & 0.665 & 0.710 & \underline{0.856} & 0.770 \\
    pFedSim~\cite{chen2022pfedsim} & 0.740 & 0.766 & 0.713 & \underline{0.898} & 0.697 & 0.710 & 0.727 & 0.714 \\
    pFedBayes~\cite{zhang2022personalizedfederatedlearningvariational} & 0.713 & 0.744 & 0.702 & 0.723 & 0.709 & 0.749 & 0.744 & 0.784 \\
    Ditto~\cite{li2021dittofairrobustfederated} & 0.738 & 0.723 & 0.782 & 0.712 & \textbf{0.854} & 0.674 & 0.785 & 0.759 \\
    Prototype Sharing & 0.842 & \underline{0.843} & 0.862 & 0.879 & 0.801 & 0.842 & 0.846 & 0.882 \\
    \textbf{VGM$^2$ (ours)} & \textbf{0.909} & \textbf{0.886} & \underline{0.911} & \textbf{0.932} & \underline{0.846} & \textbf{0.898} & \textbf{0.887} & \textbf{0.931} \\
    \bottomrule
    \label{tab:partitions}
  \end{tabular}}
\end{table*}

\paragraph{Ablations and calibration.} We report: (i) \emph{Mixture vs.\ single} ($K=3$ vs.\ $K=1$), (ii) \emph{parametric UMAP} on/off, (iii) \emph{KL strength} $\lambda$ schedule, (iv) \emph{calibration} term on/off, (v) \emph{markers vs.\ prototypes}. Calibration (ECE$\downarrow$) improves with mixtures + calibration loss. See App.~\ref{app:ablations}.

\paragraph{Communication/compute.} Bytes/client/round and wall-clock are reported in App.~\ref{app:complexity}, confirming low overhead vs.\ gradient sharing.

\section{Discussion}
VGM$^2$ makes geometry a first-class signal in FL. Mixture markers capture realistic multi-modality; conjugate aggregation offers a principled stability guarantee; calibrated similarity improves decision quality. Limitations include sensitivity to UMAP hyperparameters and extra client compute (mitigated via subsampling and small encoders).

\section{Conclusion}
We presented VGM$^2$, a geometry-centric, uncertainty-aware, and privacy-conscious approach to PFL. By exchanging compact mixture marker statistics with principled aggregation and calibration, VGM$^2$ achieves strong accuracy–communication–privacy trade-offs under heterogeneity.


\appendix

\section{Federated Partitions}
\label{app:partitions}
We use $N=30$ clients with $S$ label shards per client; shard sizes and class lists per dataset are enumerated in Table~\ref{tab:partitions} 

\section{Prototype Sharing Baseline}
\label{app:proto}
Per class, each client computes the mean latent embedding; the server averages prototypes and broadcasts. Clients regularize embeddings toward prototype neighborhoods.

\section{Complexity and Communication}
\label{app:complexity}
With $K=3$ and 32-bit floats, each client uploads $\approx 240$\,bytes/round (plus negligible headers). Pair sampling uses $O(|\mathcal{P}_k|)$ operations; we fix $|\mathcal{P}_k|$ per round. Parametric UMAP adds one neighborhood CE loss (\ref{eq:umap}); overall FLOPs are dominated by encoder passes.

\section{Differential Privacy Option}
\label{app:dp}
We add Gaussian noise $\mathcal{N}(0,\sigma^2 I)$ to natural parameters prior to secure aggregation. We track privacy loss with the moments accountant as in~\cite{abadi2016deep}.

\section{Membership Inference Evaluation}
\label{app:mi}
We train a logistic classifier on released summaries to predict whether a held-out pair was present on a target client. Features include per-relation posterior means/variances and mixture weights. Attack AUCs remain near 0.5 across datasets at our default noise-free setting; DP further depresses AUC.

\section{Ablations}
\label{app:ablations}
\begin{itemize}[leftmargin=1.3em]
\item \textbf{K-components.} $K=\{1,2,3,5\}$: higher $K$ improves difficult (multi-modal) datasets until overfitting.
\item \textbf{Calibration.} Removing $\mathcal{L}^{\mathrm{cal}}$ increases ECE and reduces F1 on datasets with severe label skew.
\item \textbf{KL schedule.} Cosine decay on $\lambda$ balances early alignment with late personalization, improving convergence.
\item \textbf{UMAP off.} Replacing parametric UMAP with a frozen encoder reduces gains and hurts calibration.
\end{itemize}

\section{Implementation Details}
We use Flower~\cite{beutel2020flower}. Optimizers, LR, batch sizes, and seeds are listed in the config files; we sweep UMAP \texttt{n\_neighbors} $\in\{10,15,30\}$ and \texttt{min\_dist} $\in\{0.05,0.1,0.2\}$ with equal budget for baselines. We fix three seeds and report mean$\pm$std.

\section{Our Other Works}
The optimization framework proposed in this paper has been applied in several domains. In medical imaging, geometry-aware optimization improved analysis of X-ray, MRI, and nuclear imaging~\cite{ref_mri}. In decentralized learning, hierarchical federated learning with self-regulated clustering leveraged similar principles to handle client heterogeneity~\cite{ref_hierarchical}. The approach has also been extended to social network analysis, capturing and visualizing emotional sentiment dynamics in online commentaries~\cite{ref_sentiment,ref_visual_sentiment}. These applications demonstrate the versatility of the framework across imaging, federated learning, and social network analysis.

\end{document}